\documentclass[conference]{IEEEtran}
\IEEEoverridecommandlockouts

\usepackage{cite}
\usepackage{amsmath,amssymb,amsfonts}
\usepackage{algorithmic}
\usepackage{graphicx}
\usepackage{textcomp}
\usepackage{xcolor}
\usepackage{bbold}
\usepackage{multirow, multicol} %
\usepackage[caption=false,font=footnotesize]{subfig} %
\usepackage{url} %
\def\BibTeX{{\rm B\kern-.05em{\sc i\kern-.025em b}\kern-.08em
    T\kern-.1667em\lower.7ex\hbox{E}\kern-.125emX}}
\begin{document}

\title{CenterDisks: Real-time instance segmentation with disk covering\\
\thanks{We acknowledge the support of the Natural Sciences and Engineering Research Council of Canada (NSERC), the Institut for data valorisation (IVADO) and the Apogée fund.}
}

\author{\IEEEauthorblockN{Katia Jodogne-del Litto}
\IEEEauthorblockA{\textit{LITIV Lab., Polytechnique Montréal} \\
Montréal, Canada \\
katia.jodogne-del-litto@polymtl.ca}
\and
\IEEEauthorblockN{Guillaume-Alexandre Bilodeau}
\IEEEauthorblockA{\textit{LITIV Lab., Polytechnique Montréal} \\
Montréal, Canada \\
gabilodeau@polymtl.ca}
}
\maketitle

\begin{abstract}
Increasing the accuracy of instance segmentation methods is often done at the expense of speed. Using coarser representations, we can reduce the number of parameters and thus obtain real-time masks.

In this paper, we take inspiration from the set cover problem to predict mask approximations. Given ground-truth binary masks of objects of interest as training input, our method learns to predict the approximate coverage of these objects by disks without supervision on their location or radius. Each object is represented by a fixed number of disks with different radii. In the learning phase, we consider the radius as proportional to a standard deviation in order to compute the error to propagate on a set of two-dimensional Gaussian functions rather than disks.

We trained and tested our instance segmentation method on challenging datasets showing dense urban settings with various road users.

Our method achieve state-of-the art results on the IDD and KITTI dataset with an inference time of 0.040 s on a single RTX 3090 GPU.

The code is available at: 

https://github.com/KatiaJDL/CenterDisks
\end{abstract}

\section{Introduction}

The development of intelligent vehicles is a major technological advance, which relies heavily on computer vision to observe and interpret the environment. The use cases are multiple in the context of light individual vehicles or public transport, and range from driving assistance mechanisms to autonomous driving. The deployment in urban areas requires the ability to detect road users in real-time and accurately in dense scenes. 
However, the deep learning models used for image processing are generally quite resource intensive. One of the factors at play is their large number of parameters. Reducing their sizes allows both to speed them up and to require less computing resources. Smaller models can be used with less powerful and lighter hardware, for example in embedded systems. The climate crisis and its challenges also make it necessary to be aware of the resources consumed for training and production.

To localize precisely road users, detections with bounding boxes are not enough. We need to segment the image to get the pixels belonging to each object of interest. But predicting binary masks is quite slow, and most precise instance segmentation do not reach real-time performances \cite{he_mask_2017, liu_path_2018}. One way to achieve a trade-off between speed and accuracy is to use mask approximations, with a lighter representation and therefore less parameters.

\begin{figure}[t]
\centering
\subfloat[Original image.]{\includegraphics[width=0.22\textwidth]{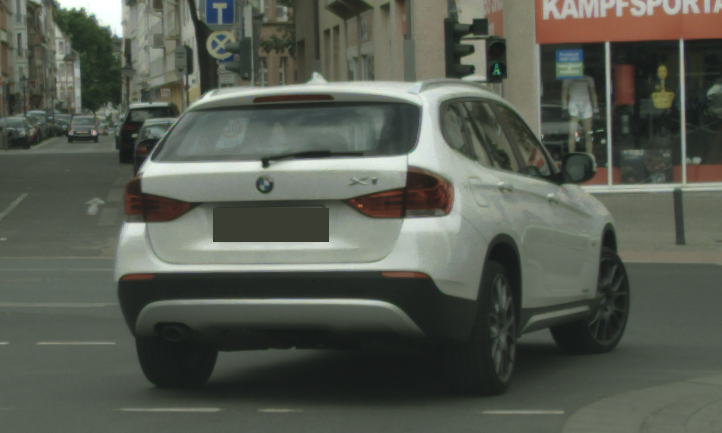}%
\label{fig:car}}
\hfill
\subfloat[Binary mask.]{\includegraphics[width=0.22\textwidth]{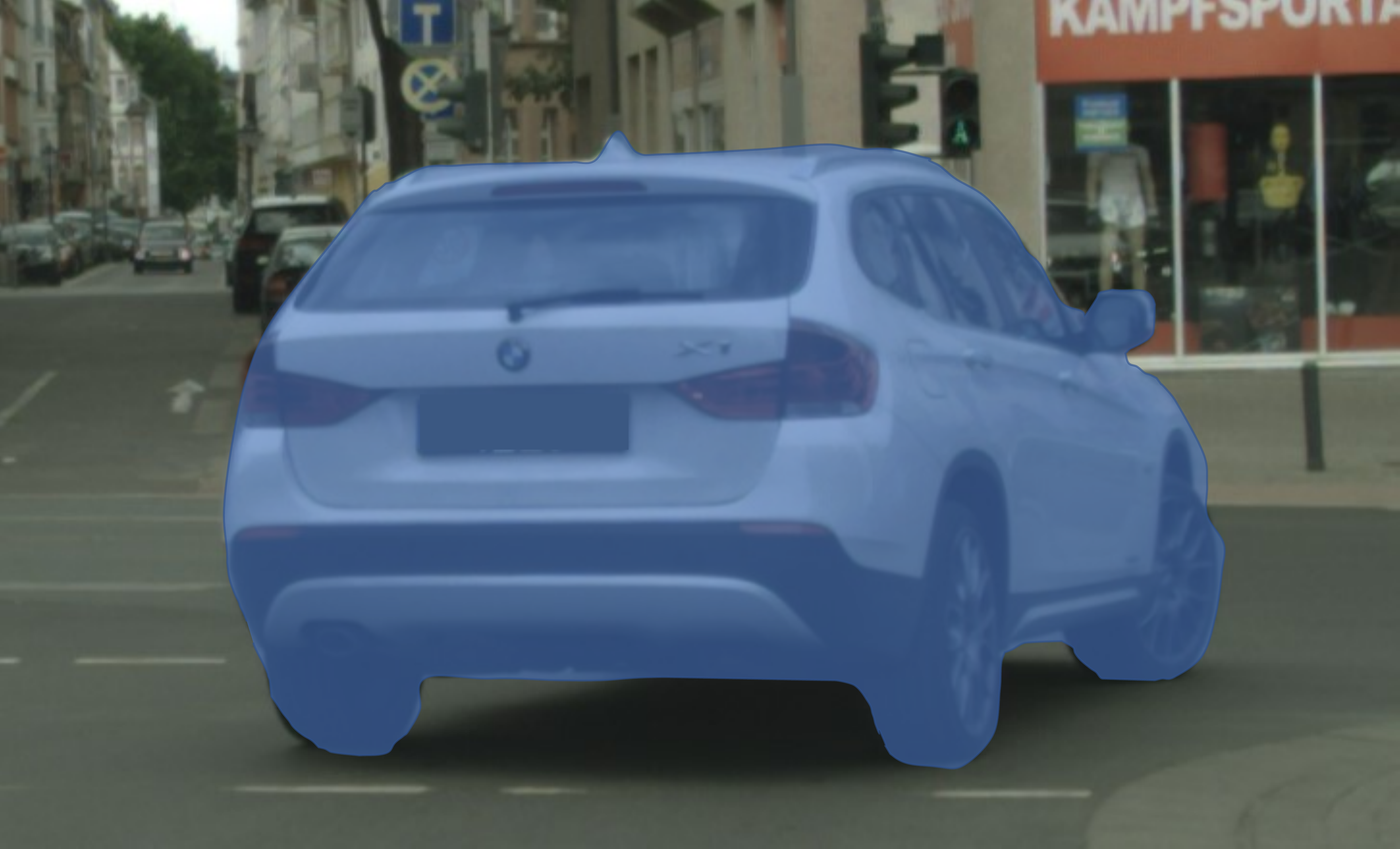}%
\label{fig:car_mask}}
\hfill
\subfloat[Set of covering disks.]{\includegraphics[width=0.22\textwidth]{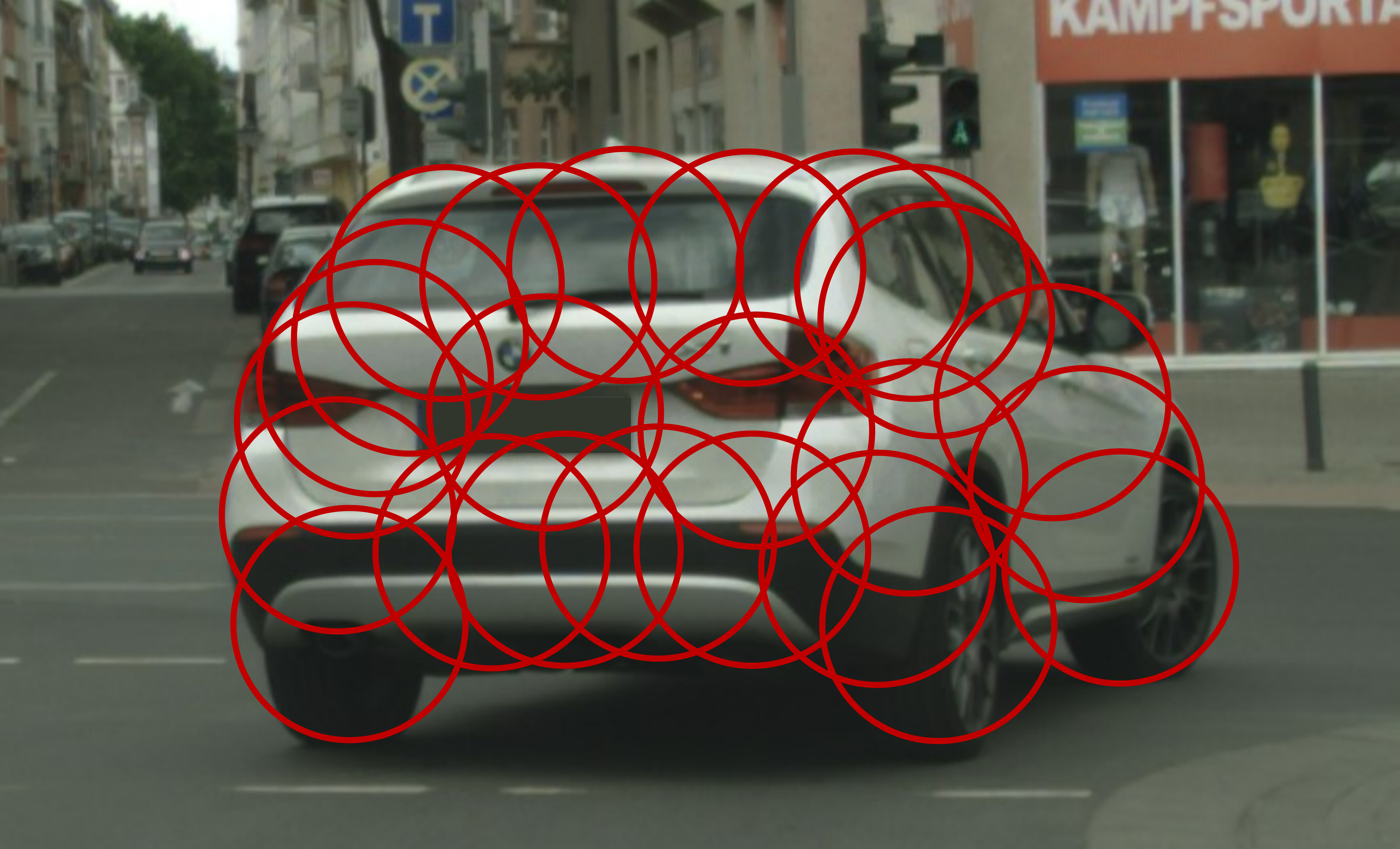}%
\label{fig:car_disks}}
\hfill
\subfloat[Mask produced by the disks.]{\includegraphics[width=0.22\textwidth]{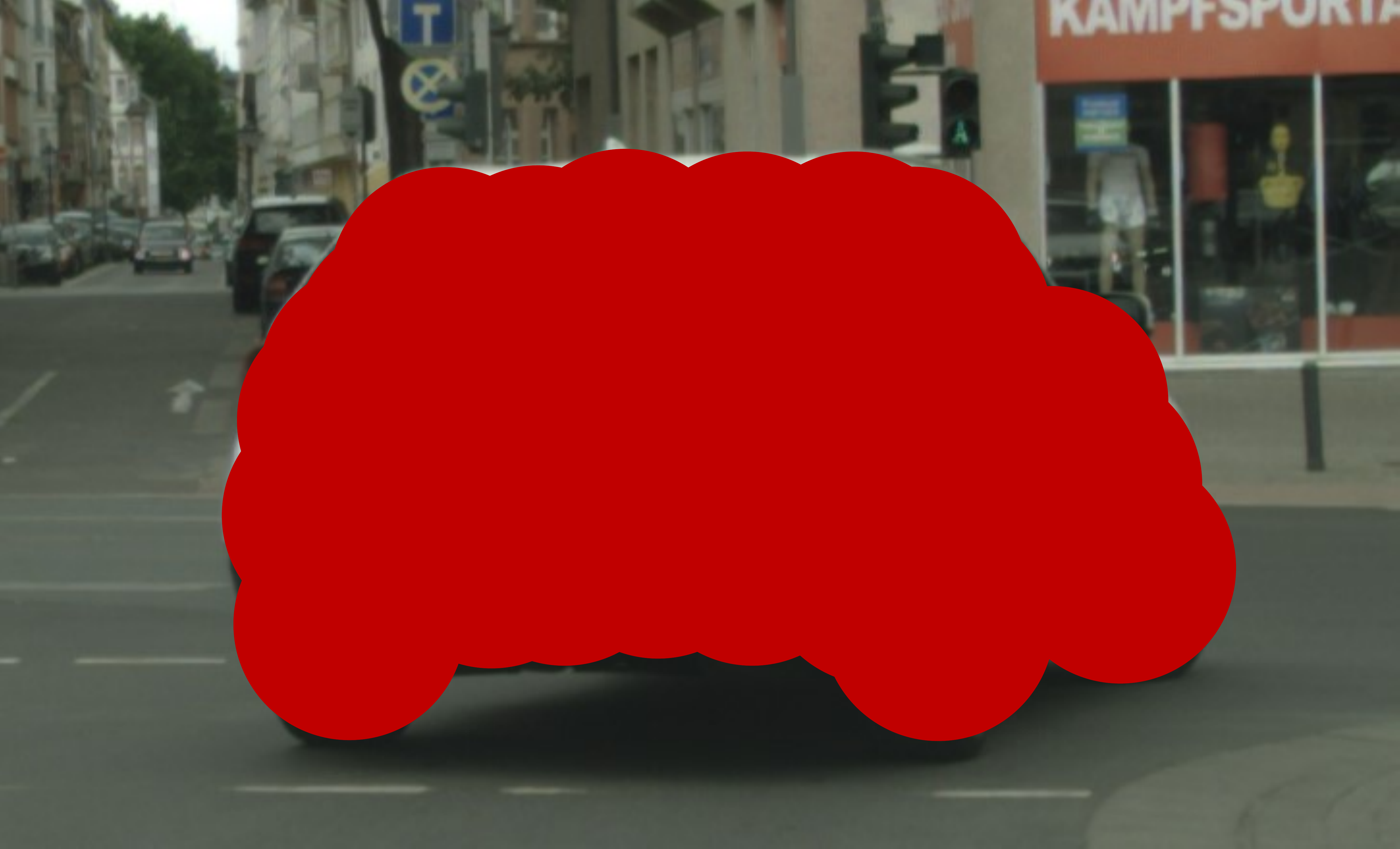}%
\label{fig:car_disksmask}}
\caption{Disk covering on a Cityscapes image crop. The set of covering disks is composed of 24 disks with identical radius.}
\label{fig:masks_comparison}
\end{figure}

One way to simplify a binary mask is to use only its contours \cite{xie_polarmask_2020, perreault_centerpoly_2021, hurtik_poly-yolo_2022, peng_deep_2020}, but we can also be interested in the possibility of approximating a binary mask with few parameters by representing the interior of the mask directly. This can be viewed as a mask covering problem. We no longer characterize each pixel of the image, but we want to project shapes on it. The simplest and least expensive way in terms of number of parameters to represent a surface is a disk. Only three parameters are needed, the two coordinates of the center and the radius. Thus, we propose to approximate a complex binary mask with a set of disks (see Figure \ref{fig:masks_comparison}).

One of the main advantages of this representation is that it does not require direct supervision or a custom ground-truth. There is no need for a ground-truth representing the centers and radii of the sets of disks. The optimization is performed directly on the binary masks. 

\clearpage

Our contributions are the following:
\begin{itemize}
    \item We propose a new approach for mask approximation with disk covering;
    \item We show that a deep neural network can learn to cover a surface by positioning a set of disks;
    \item Our proposed method is real-time and achieves state-of-the art results on the IDD and KITTI test set.
\end{itemize}

\section{Related works}

\paragraph{Real-time instance segmentation}
Most traditional instance segmentation methods, like Mask R-CNN \cite{he_mask_2017}, PANET \cite{liu_path_2018}, or more recently Sam \cite{kirillov_segment_2023}, cannot reach real-time performances. But techniques were developed to speed them up. First, we can mention sparse representation of object localization. In SOLOv2~\cite{wang_solov2_2020}, the images are divided into cells, and each cell corresponds to a mask, representing the potential object which center lies in this cell. SparseInst \cite{cheng_sparse_2022} also uses sparse feature maps to produce mask kernels but does not need to localize the objects by their center. Spatial Sampling Net \cite{mazzini_spatial_2019} produces a non-uniform density map. It follows the object distribution, and the masks are obtained with a diffusion process through a spatial sampling operator.

Furthermore, combining bounding boxes and segmentation masks can be an acceleration factor. ESE SEG \cite{xu_explicit_2019} performs at the same time the bounding box prediction and the object segmentation, and Box2Pix \cite{uhrig_box2pix_2018} combines semantic segmentation and bounding boxes. Finally, real-time can be achieve with a simplified representation of masks, with a combination of mask prototypes \cite{bolya_yolact_2019}, or with polygonal mask approximations \cite{hurtik_poly-yolo_2022, perreault_centerpoly_2021}.

\begin{figure*}[ht]
    \centering
    \includegraphics[width=0.7\textwidth]{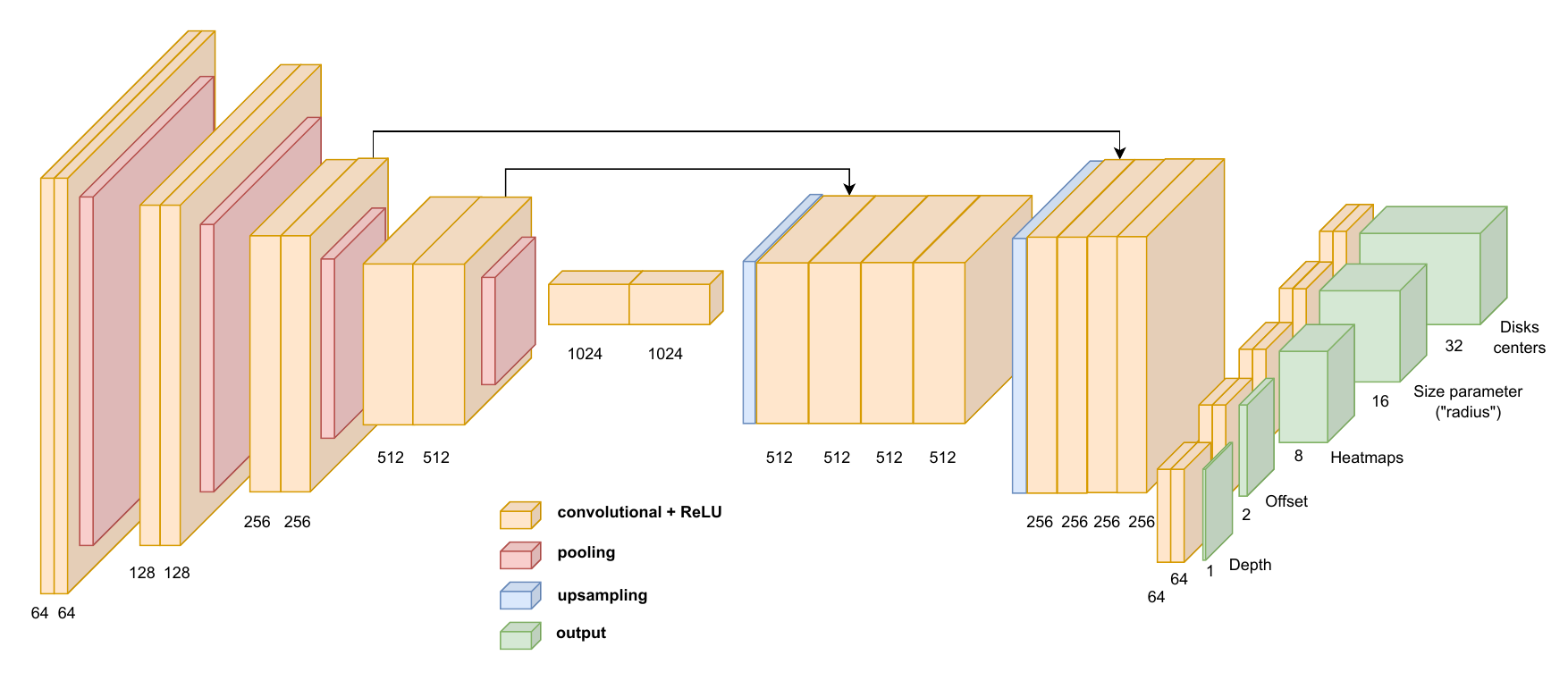}
    \caption{Architecture for CenterDisks. The backbone is represented here as an Houglass backbone \cite{newell_stacked_2016}. The five heads predict the heatmap for object centers, the offsets from this center, the relative depths, and the parameters of the disk sets. The number of parameters displayed on this figure is only as an illustration. For implementation details, please refer to section \ref{sec:implementation} and to the code provided.}
    \label{fig:architecture}
\end{figure*}

\paragraph{Instance segmentation with mask approximations}
Except for YOLACT \cite{bolya_yolact_2019}, which uses mask prototypes and predicted coefficients to combine them, most mask approximation methods consider only the boundary of the objects. A popular representation is the polygonal mask.

Polygon-RNN \cite{castrejon_annotating_2017} and Polygon-RNN++ \cite{acuna_efficient_2018} use a recurrent neural network to determine the next vertex. These methods were meant for semi-automated annotation systems, and are really slow. PolarMask \cite{xie_polarmask_2020} fixes the angles of the vertices in a polar representation, which produce a star structure. Poly-YOLO \cite{hurtik_poly-yolo_2022} also uses a polar grid, with free angles and a dynamic number of vertices, chosen with a dedicated confidence score during inference phase. CenterPoly \cite{perreault_centerpoly_2021} and CenterPolyV2 \cite{jodogne-del_litto_real-time_2023} generate simultaneously heatmaps for object detection, and polygon vertices for each pixel. 
Nevertheless, most of these methods do not consider potential holes in the masks.

However, non-polygonal representations exist. ESE SEG \cite{xu_explicit_2019} uses an approximation of object boundaries based on Chebyshev polynomials as mask approximation. 
FourierNet \cite{riaz_fouriernet_2021} and SCR \cite{bahl_scr_2022} use instead Fourier series and add to the output a differentiable decoder that allows learning directly on the final shape of the masks. DeepSnake \cite{peng_deep_2020} directly generates the contours and then iteratively distorts them to get closer to the actual contours of the objects. BCondInst \cite{zhang_boundary-preserving_2022} and BshapeNet \cite{kang_bshapenet_2020} use boundary predictions to improve its mask predictions.

\paragraph{Set cover problem}
The mask approximation methods presented above rely on the detection of the object contour. One can also consider a representation based on the inside of the objects. Approximating a complex surface with a set of shapes can be seen as a sub-problem of the general set cover problem, which is NP-complete \cite{karp_reducibility_1972}.
Depending on the evaluation criteria for the covering, there are several mathematical results to this computational geometry problem for simple configurations.

The problem of covering the unit disk uses an indirect approach by requiring the minimization of the common radius of a set of disks to cover the whole surface. This mathematical problem is solved exactly for different values of $n$: 1, 2, 3, 4, 5, 7 \cite{neville_solution_1915}. 
and approximate radius values have been obtained up to $n=10$ \cite{zahn_black_1962}. Here the location of the disks is secondary, and the radius is the value to be minimized.
For less regular surfaces, under real-life conditions, this problem can be interpreted as the placement of radio antennas (or base station placement) \cite{salhieh_power_2001}. There are approximation algorithms for convex polygons \cite{xu_connected_2018}. For more complex surfaces, the optimization algorithms range from linear integer programming \cite{horster_approximating_2006}, greedy \cite{munishwar_coverage_2013}, to meta-heuristic like genetic algorithms \cite{han_genetic_2001}.

Thus, there is no exact algorithm to find the position of disks covering any surface by minimizing their radii. The existing algorithms use optimization techniques. In our case, the input is an image with multiple areas that should be covered as exactly as possible. It is not a known surface. Nevertheless, we postulate and demonstrate with practical results that it is possible to optimize the overlap by deep learning with the complete mask as the only supervision.

\section{Method}

Our method is based on the object detector CenterNet \cite{zhou_objects_2019}, which locates objects by their center and regress from it the coordinates of their bounding box. Instead of bounding boxes, we adapt the prediction heads to obtain a set of centers and radii for each object detected. The architecture of our method is shown in the figure \ref{fig:architecture}. There are five prediction heads. The heatmap and offset heads come directly from CenterNet \cite{zhou_objects_2019}. With one heatmap for each semantic category, we can extract the peaks, which correspond to the center of the objects. The relative depth head is inspired by CenterPoly \cite{perreault_centerpoly_2021}. We added two heads for centers and radii prediction.

\subsection{Gaussian projection}

The covering is densely predicted: A set of $N$ disks is predicted for each pixel, and only the ones corresponding to a peak of the heatmap are kept. 
Therefore, a set of $N$ disks covers one object. That represents $2N$ coordinates for the center of the disks $x_1, y_1, x_2, y_2, ..., x_n, y_n$ and $M \leq N$ size parameters $\sigma_1, ... \sigma_M$. These last values can be considered as the standard deviation of a Gaussian function in two dimensions. 

Each center is matched with a standard deviation using an association function $\lambda$. $N, M, \lambda$ are hyper-parameters of the model. If $z$ is the index of a center, $\lambda(z)$ is the index of the corresponding radius. For example, if the standard deviation is the same for all centers, we must fix $M=1$ and $\lambda: z \rightarrow 1$. For each center to get its personalized standard deviation, $M=N$ and $\lambda: z \rightarrow z$. We performed an ablation study to choose the proportion of size parameters compared to the number of centers $N$.

The equation for the Gaussian function of center $i$ is

\begin{equation}    
    f_i (x, y) = e^{- \frac{(x-x_i)^2+(y-y_i)^2}{2\sigma_{\lambda(i)}^2}},
\end{equation}

with which we can deduce the global probability map for one object:

\begin{equation}
    f (x, y) = \sum_{i=0}^{N} e^{- \frac{(x-x_i)^2+(y-y_i)^2}{2\sigma_{\lambda(i)}^2}}.
    \label{eq:proba_map}
\end{equation}

The final result of the prediction head for the disks is a probability map. 

\subsection{Training}

Since the output is a sum of Gaussian functions, we have to apply a normalization function on these probability maps to reduce their range to $[0,1]$. This allows us to use classic loss functions for segmentation tasks. We chose to use the hyperbolic tangent function. 

No custom ground-truth (GT) is necessary. We can simply use the binary masks provided with the datasets. The loss function used for the training of the model is defined as follow:
\begin{multline}
    Loss = W_{hm} Loss_{hm} + W_{offset} Loss_{offset} \\
    + W_{depth} Loss_{depth} + W_{disks} Loss_{disks},
    \label{eq:global_loss}
\end{multline}

where $W_{hm}$, $W_{offset}$, $W_{depth}$ are the respective weights of the loss for the prediction heads for heatmap, offset and relative depth, and  $W_{disks}$ is the weight for the loss for the disk heads (head for disk centers and head for standard deviations). The loss for the heatmaps describing the center of objects is the focal loss \cite{lin_focal_2017}, and $Loss_{offset}$ and $Loss_{depth}$ are L1 loss functions.

The loss function used for the training of the two disk heads should compute the difference between a GT binary mask and a probability map. We propose the use of two possible loss functions: binary cross-entropy \cite{yi-de_automated_2004} and Dice loss \cite{milletari_v-net_2016}. The first one is a local loss, based on the distribution, while the second is a regional loss, taking into account the whole image. For a pixel $k$ with its coordinates $(x,y)$, $z_k \in \{0,1\}$ is the GT value and $p_k=f(x,y) \in [0,1]$ is the predicted value, corresponding to \ref{eq:proba_map}. The losses are defined by

\begin{equation}
    L_{cross\_entropy} = - \frac{1}{N} \sum_{k} z_k log(p(p_k)) + (1-z_k) log(1 - p_k)
\end{equation}
and
\begin{equation}
    L_{Dice} = 1 - \frac{2 \sum_{i=1}^N z_k p_k}{\sum_{k} z_k + \sum_{k} p_k + \epsilon},
\end{equation}

$\epsilon$ being a smoothing term. $Loss_{disks}$ is then one of those two.


\subsection{Inference}
\label{sec:postprocessing}

\begin{figure}
\centering
\subfloat[Mask generated with superposition of Gaussian functions.]{\includegraphics[width=0.24\textwidth]{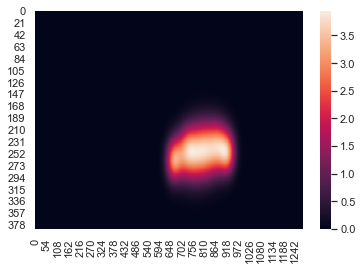}%
\label{fig:heatmap}}
\hfill
\subfloat[Final binary mask.]{\includegraphics[width=0.24\textwidth]{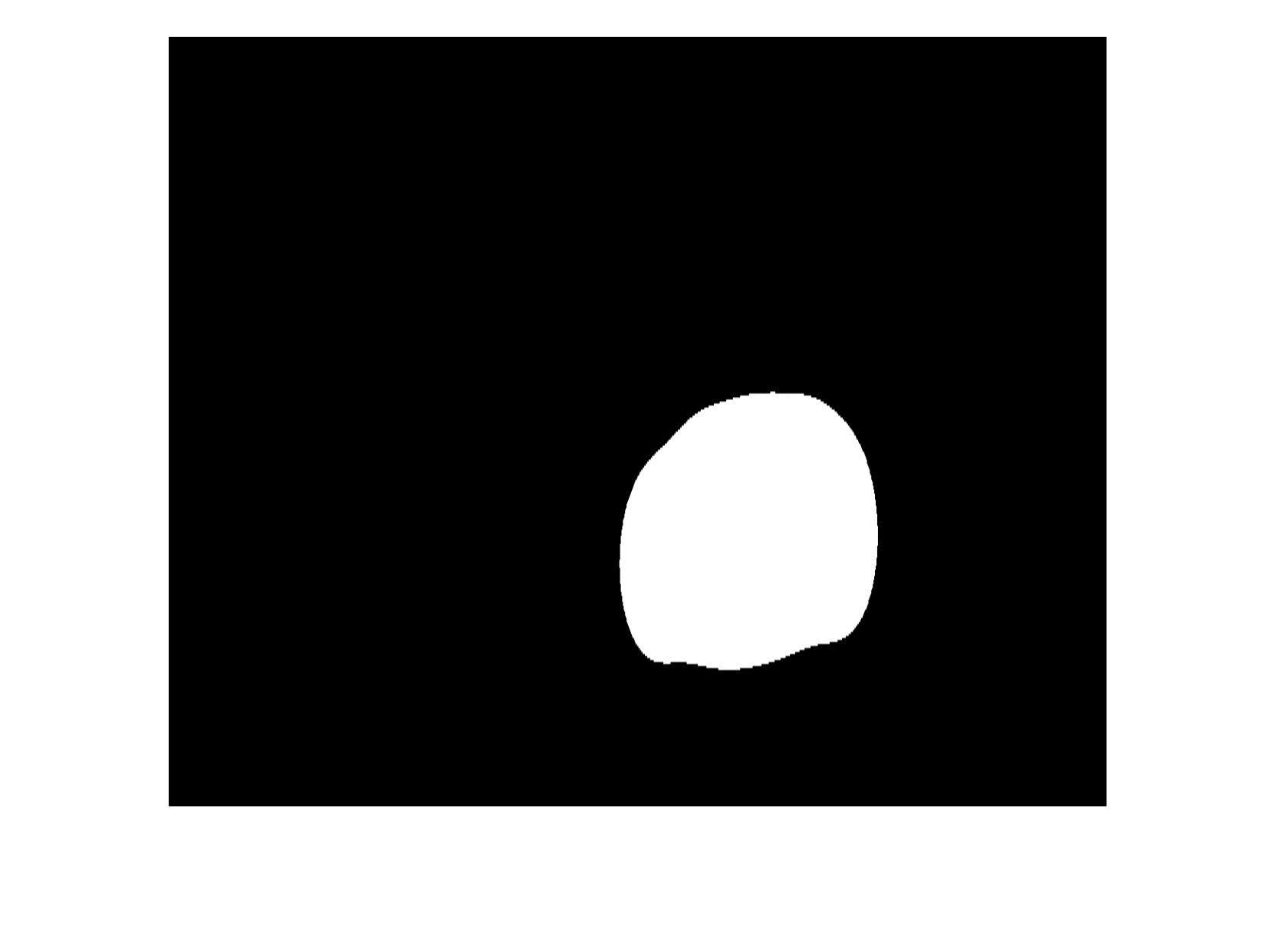}%
\label{fig:binary}}
\caption{Thresholding with $\alpha = 0.5$ on the predicted set of disks at inference phase.}
\label{fig:tresholding}
\end{figure}

In the inference phase, we perform no normalization on the probability maps representing the objects. After projection, with two-dimensional Gaussian functions, the predicted binary mask is selected with a threshold $\alpha$ (Figure \ref{fig:tresholding}).

We may perform contour smoothing with the Douglas-Peucker algorithm \cite{douglas_algorithms_1973}. With simple computer graphics tools, we retrieve a polygon approximation of the contour from the binary mask \cite{suzuki_topological_1985} and perform a reduction of the number of vertices. The hyper-parameter $\beta$, used as a maximum distance to the original outline, is proportional to the initial perimeter. This post-processing step can reduce the roundness of the mask, which is inherent to disks.


\section{Experiments}

\subsection{Implementation}
\label{sec:implementation}

The focus of our study is the segmentation of road users in urban settings. We used the Cityscapes \cite{cordts_cityscapes_2016}, IDD \cite{varma_idd_2019} and KITTI datasets \cite{geiger_are_2012}. We performed ablation studies on the Cityscapes validation set.
The instance categories used correspond to road users: bicycle, bus, car, motorcycle, person, rider, train, truck. The data splits used for training, validation and testing are predefined for all datasets. The images of the Cityscapes dataset are recorded in different cities in Germany. It includes 5,000 images, with a standard resolution of $2048 \times 1024$. In the Indian Driving Dataset (IDD), there are approximately 10,000 images, with sizes varying from $1920 \times 1080$ to $1280 \times 964$. Finally, KITTI dataset includes 400 images for segmentation, with a size of $1280 \times 384$. 

Our main evaluation metric is the Average Precision (AP) \cite{lin_microsoft_2014}. It is an aggregation of the average precision for different IoU thresholds, from 50\% to 95\%. The AP50\%, for average precision with minimum IoU of 50\%, and AP50m and AP100m, for objects within a range of 50m and 100m respectively, are also used.
Moreover, for all these metrics, we can have access to the average over all categories or to the detailed results.

We implemented our method with Pytorch \cite{paszke_pytorch_2019}. We use the Hourglass network \cite{newell_stacked_2016}, with one stack as a backbone for all our experiments. The backbone, heatmap head, and offset head are pre-trained on COCO \cite{lin_microsoft_2014}. We first trained on Cityscapes and then fine-tuned our model for KITTI and IDD. For training, we used classical data augmentation techniques, with a resolution of $1024 \times 512$: color augmentation, random cropping, and flipping. The loss weights are $W_{hm} = 1$, $W_{disks} = 1$, $W_{depth} = 0.1$ and $W_{offset} = 0.1$. For the disk heads we used the Dice loss. 

The model was trained for 240 epochs with a batch size of 4 on a single RTX 3090 GPU with the adam optimizer \cite{kingma_adam_2015}. The starting learning rate used was 2e-4, and we divided the learning rate by ten at epochs 90 and 120. The threshold for the inference was fixed to $\alpha = 0.5$. We use 16 disks with 16 different radii.

\begin{figure}[t]
    \centering
    \subfloat[AP and AP50\% for each semantic category.]{\includegraphics[width=0.2\textwidth]{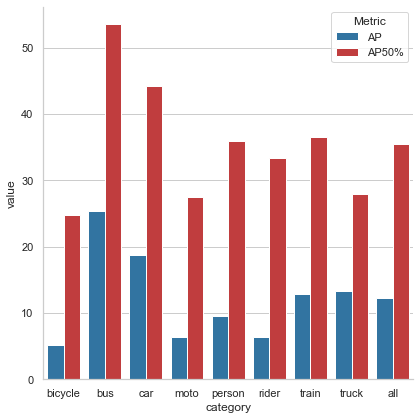}%
    \label{fig:bars_categories}}
    \hfill
    \subfloat[Frequency of each category in the Cityscapes dataset.]{\includegraphics[width=0.2\textwidth]{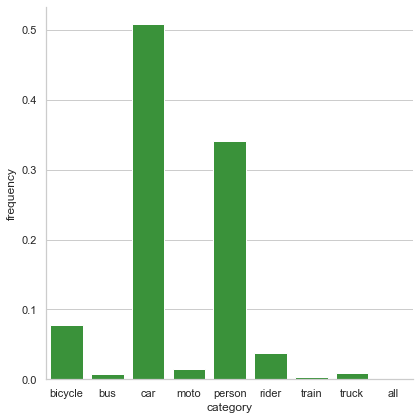}%
    \label{fig:bars_frequencies}}
    \caption{Results on the Cityscapes validation set by categories.}
    \label{fig:results_categories}
\end{figure}

\subsection{Results}

\begin{table*}[th]
\caption{Results on the Cityscapes test set. If the runtimes were not explicitly stated in the original paper, there are estimated based on our knowledge of the method.
Results are taken from the original papers or public online benchmarks, unless stated otherwise. \textbf{Boldface: Best results for real-time methods}. \underline{Underline: Best results overall}.}
  \label{results_cityscapes}
  \centering
  \begin{tabular}{llrrrrr}
    \hline
    Method & Mask type & AP \textuparrow & AP50\% \textuparrow & AP100m \textuparrow & AP50m \textuparrow & Runtime (s) \textdownarrow\\
    \hline
    Mask-RCNN \cite{he_mask_2017} & Full & 26.22 & 49.89  & 37.63 & 40.11 & $\simeq$ 0.2 \\
    PANET \cite{liu_path_2018} & Full & 31.80 & 57.10 & 44.20 & 46.00 & $>$ 1 \\
    PolyTransform \cite{liang_polytransform_2020} & Polygon & \underline{40.10} & \underline{65.90} & \underline{54.80} & \underline{58.00} & $>$ 1 \\
    DeepSnake \cite{peng_deep_2020} & Outline & 31.70 & 58.40 & 43.20 & 44.70 & 4.6 \\
    Polygon-RNN++ \cite{acuna_efficient_2018} & Polygon & 25.50 & 45.50 & 39.30 & 43.40 & $>$ 1\\ 
    \hline
    Spatial Sampling Net \cite{mazzini_spatial_2019} & Full & 9.20 & 16.80	& 16.4 & 21.4 & \textbf{0.009} \\
    Box2Pix \cite{uhrig_box2pix_2018} & Full & 13.10 & 27.20 & - & - & 0.092 \\
    Poly-YOLO \cite{hurtik_poly-yolo_2022} & Polygon & 11.50 & 26.70 & - & - & 0.049 \\
    Poly-YOLO lite \cite{hurtik_poly-yolo_2022} & Polygon & 10.10 & 23.90 & - & - & 0.027 \\
    CenterPoly \cite{perreault_centerpoly_2021} & Polygon & \textbf{15.54} & \textbf{39.49} & \textbf{23.33} & \textbf{24.45} & 0.045 \\
    \hline
    CenterDisks (Ours) & Shape set & 7.36 & 25.89 & 11.64 & 12.37 & 0.040 \\
    \hline
  \end{tabular}
\end{table*}

The results for our method on the test sets of Cityscapes, IDD and KITTI are presented in the Table \ref{results_cityscapes}, Table \ref{results_idd} and Table \ref{results_kitti}. We included methods with real-time performances as well as slower but more precise methods.

For the Cityscapes dataset (Table \ref{results_cityscapes}), the inference time for one image is about 0.040 second. It means that our method can be used in real-time settings. Our method reaches 7.36 for Average Precision and 25.89 for AP50\%. Except for CenterPoly, our results for AP50\% are competitive with other real-time method. The evaluation metrics are very demanding, and a low score, especially for the AP metric, does not mean that results are not usable. 

As we can see in Figure \ref{fig:results_categories}, the best performances are met on the categories that are most present in the dataset, and that have the biggest surface in average. The cars, representing almost half of the objects in the Cityscapes dataset, are especially well segmented. The shape of the objects has also a high importance, since disks are not very flexible. Compact blocks (cars, buses, trucks, trains) are easier to approximate here, whereas the handlebars of a bike or the slender shape of pedestrian are harder to get right.

However, on the Indian Driving Dataset (Table \ref{results_idd}) and on the KITTI dataset (Table \ref{results_kitti}) CenterDisks performs better than the previous methods. We reach 20.30 in AP and 49.90 in AP50\% for IDD, and for KITTI, we improve the state-of-the-art method by 3 points for the AP metric and 10.5 points for the AP50\% metric. These datasets contain less pedestrians, and no motorcycle, bicycle or rider for KITTI. We have discussed previously that the most frequently represented and the largest objects were the best segmented objects, so this can explain the very good performances on these two datasets. Moreover, IDD is more diverse inside each category, and KITTI does not have a lot of training data. Our method thus seems to have a higher generalization capacity than previous methods. 

\begin{figure*}[htbp]
    \subfloat{\includegraphics[width=0.49\textwidth]{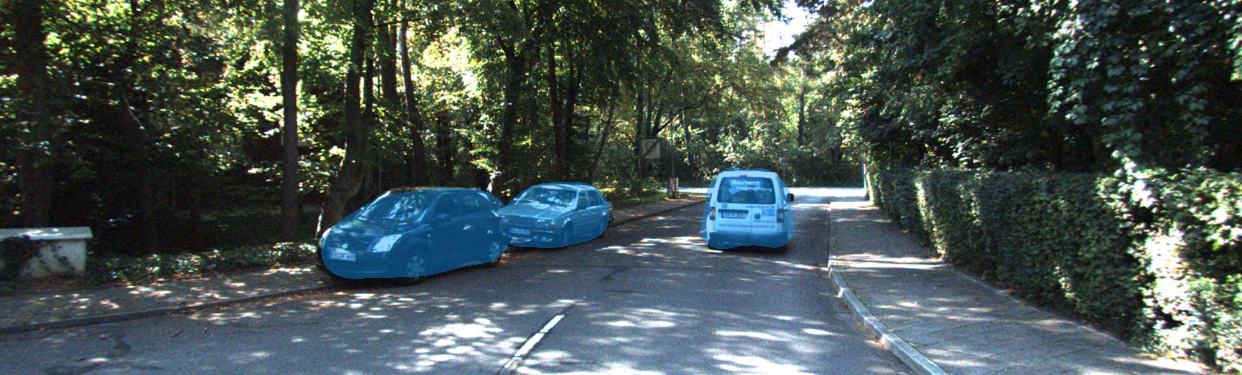}}
    \hfill
    \subfloat{\includegraphics[width=0.49\textwidth]{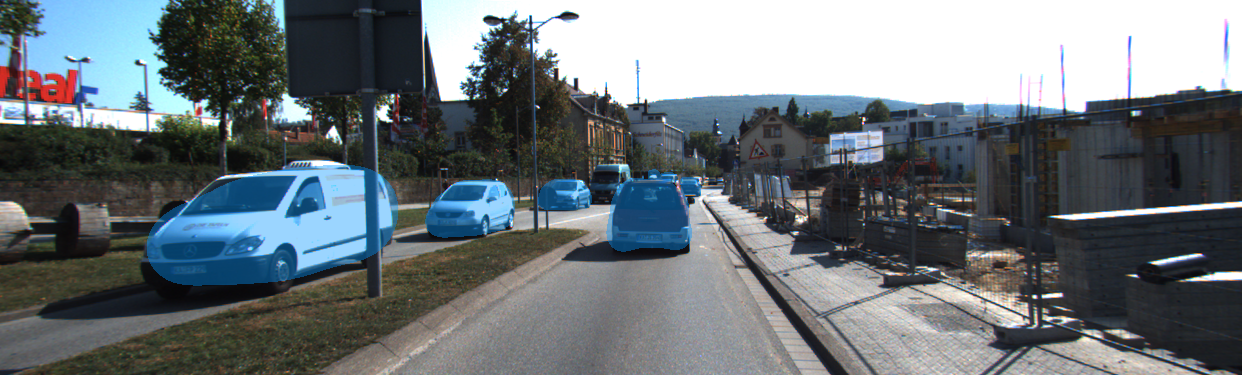}}
    \hfill
    \subfloat{\includegraphics[width=0.5\textwidth]{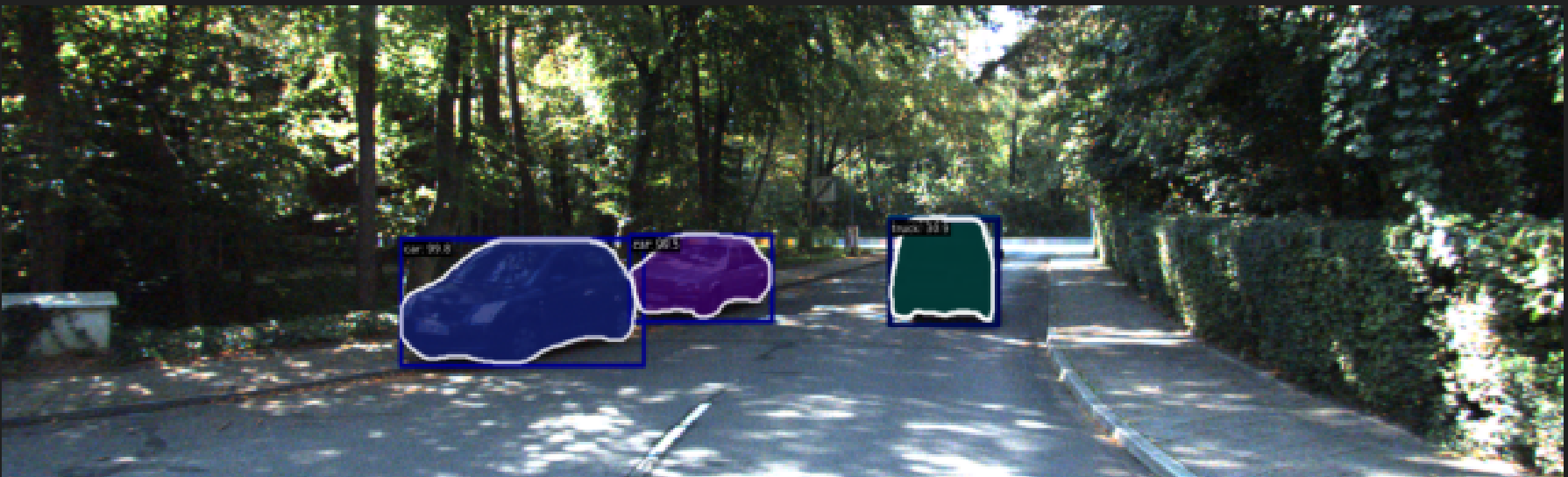}}
    \hfill
    \subfloat{\includegraphics[width=0.5\textwidth]{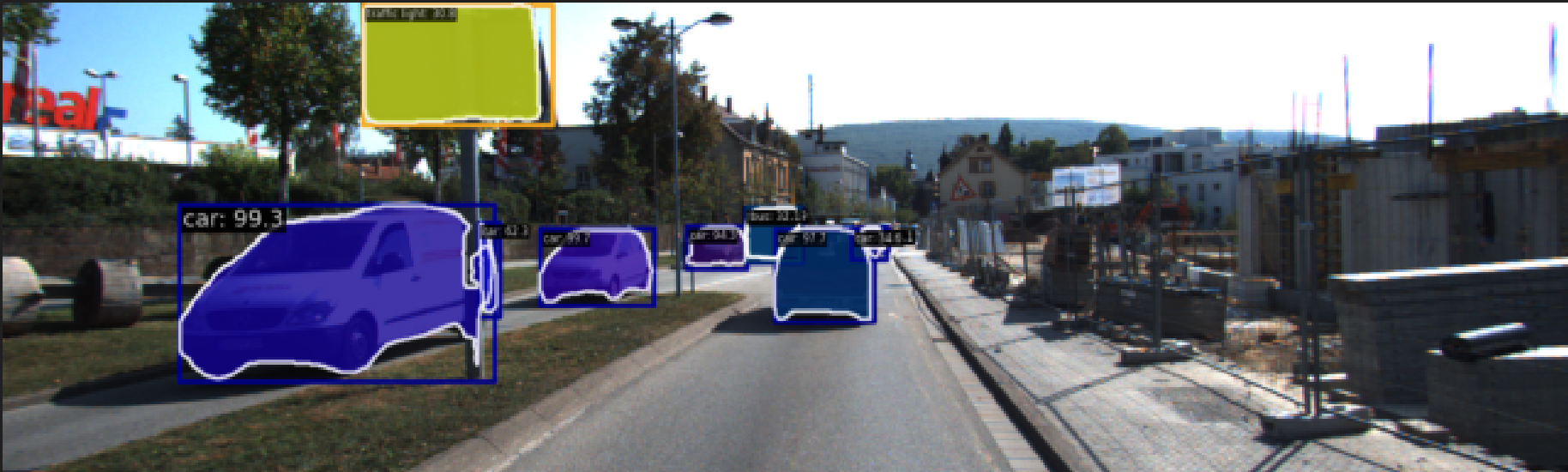}}
    \hfill
    \subfloat{\includegraphics[width=0.5\textwidth]{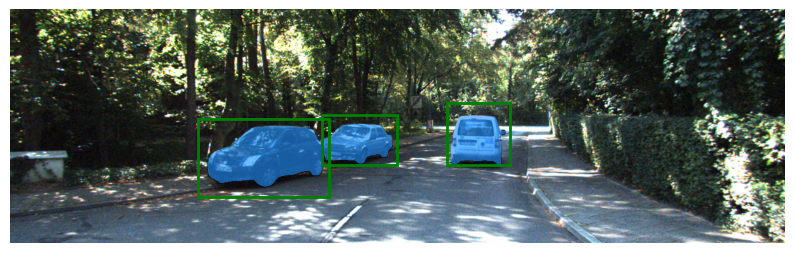}}
    \hfill
    \subfloat{\includegraphics[width=0.5\textwidth]{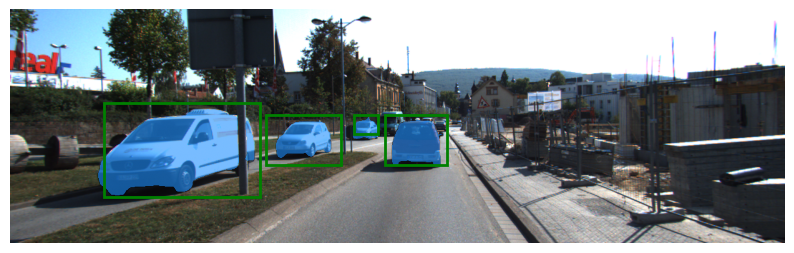}}
    \hfill
    \subfloat{\includegraphics[width=0.49\textwidth]{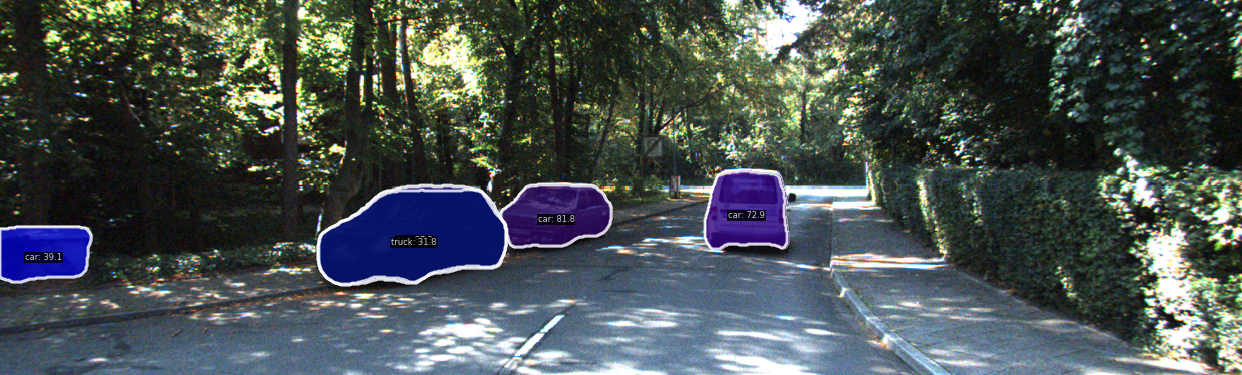}}
    \hfill
    \subfloat{\includegraphics[width=0.49\textwidth]{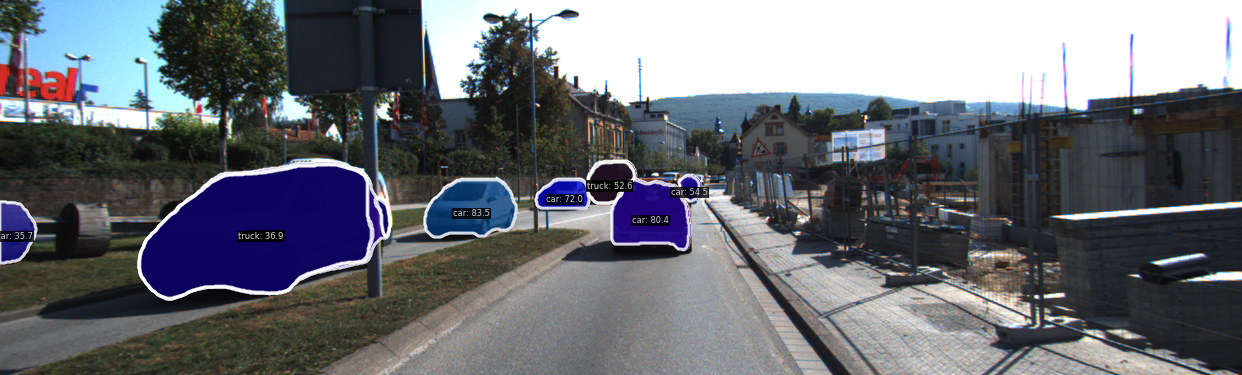}}
    \caption{Qualitative results on the KITTI test set. Comparison with state-of-the-art instance segmentation methods. From top to bottom: CenterDisks (our method), Mask-RCNN \cite{he_mask_2017}, Segment Anything \cite{kirillov_segment_2023}, SparseInst \cite{cheng_sparse_2022}. We used the pre-trained models provided by the authors, without fine-tuning them on KITTI. Best viewed on a screen.}
    \label{fig:qualitative_results_kitti}
\end{figure*}

Qualitative results for the KITTI test set and Cityscapes test set are shown on Figure \ref{fig:qualitative_results_kitti} and Figure \ref{fig:qualitative_results}. The instances are well-segmented, and some details can be captured, for example the wheels or the leg separation of pedestrians when they are big enough. The predicted masks are round-shaped. It is inherent in the structure of the mask with a set of disks, which cannot create straight lines or sharp angles. Slower methods provide more accurate masks (Figure \ref{fig:qualitative_results_kitti}), especially for details, such as the tires.

\begin{table}
\caption{Results on the IDD test set. * Results from the original IDD paper \cite{varma_idd_2019}. \textbf{Boldface: Best results for real-time methods}. \underline{Underline: Best results overall}.}
  \label{results_idd}
  \centering
  \begin{tabular}{lrrr}
    \hline
    Method & AP & AP50\% & Time (s) \\
    \hline
    Mask-RCNN \cite{he_mask_2017}* & 26.80 & 49.90 & $\simeq$ 0.2 \\
    PANET \cite{liu_path_2018}*& \underline{37.60} & \underline{66.10} & $>$ 1 \\
    \hline
    Poly-YOLO \cite{hurtik_poly-yolo_2022} & 11.50 & 26.70 & 0.049 \\
    Poly-YOLO lite \cite{hurtik_poly-yolo_2022} & 10.10 & 23.90 & \textbf{0.027} \\
    CenterPoly \cite{perreault_centerpoly_2021} & 14.40 & 36.90 & 0.045 \\
    \hline
     CenterDisks (Ours) & \textbf{20.30} &  \textbf{49.90} & 0.032 \\
    \hline
  \end{tabular}
  
\end{table}

\begin{table}
\caption{Results on the KITTI test set. \textbf{Boldface: Best results.}}
  \label{results_kitti}
  \centering
  \begin{tabular}{lllrrr}
    \hline
    Method & AP & AP50\% & Time (s)\\
    \hline
    CenterPoly \cite{perreault_centerpoly_2021}  & 8.73 & 26.74 & 0.045 \\
    \hline
    CenterDisks (Ours) & \textbf{11.75} & \textbf{37.24} & \textbf{0.033}\\
    \hline
  \end{tabular}
  
\end{table}

\section{Discussion}

\subsection{Ablation studies}

We performed ablation studies on the number of disks used for the covering (Table \ref{ablation_loss}). At first, adding more disks allow for a better covering, but then reach a plateau. The best value among the ones we tested is 16 disks. The runtime for this configuration stays also under the limit for real-time application. 

To decide between a region-based loss and a distribution-based loss, we experimented with the binary-cross entropy loss \cite{yi-de_automated_2004} and Dice loss \cite{milletari_v-net_2016}. The performance is much better using the region-based loss function Dice (Table \ref{ablation_loss}). This can be partly attributed to the imbalance between objects of interest and background, which is better taken into account by this error function.

We also checked if each disk should get its own personalized radius. The more the radii are individualized, the more accurate the covering should be. But at the same time, it can be harder to optimize more different prediction parameters. As we can see on Table \ref{ablation_variations_r}, using more different radii improves the covering. The best configuration is the one where all disks have a different radius, so it is the one we used in our other experiments.

\begin{table}
    \renewcommand{\arraystretch}{1.3}
    \centering
    \caption{Results on the Cityscapes validation set. All radius different.
    \textbf{Boldface: Best results}.}
    \label{ablation_loss}
    \begin{tabular}{lllrrr}
    \hline
    N disks & N radii & Loss & AP & AP50\% & Runtime (s)\\
    \hline
    2 & 2 & Dice & 5.03 & 19.50 & \textbf{0.032} \\
    4 & 4 & Dice & 9.51 & 33.79 & 0.036\\
    8 & 8 & Dice & 10.53 & 33.33 & 0.039\\
    16 & 16 & Dice & \textbf{12.22} & \textbf{35.47} & 0.040\\
    24 & 24 & Dice & 10.78 & 32.18 & 0.043\\
    32 & 32 & Dice & 10.65 & 33.31 & 0.046\\
    \hline
    16 & 16 & BCE & 8.00 & 27.14 & 0.040\\
    16 & 16 & Dice & \textbf{12.22} & \textbf{35.47} & 0.040\\
    \hline
    \end{tabular}
\end{table}

\begin{table}
    \renewcommand{\arraystretch}{1.3}
    \centering
    \caption{Results on the Cityscapes validation set. The loss function used for the training is the Dice loss.
    \textbf{Boldface: Best results}.}
    \label{ablation_variations_r}
    \begin{tabular}{llrrr}
    \hline
    N disks & N radii & AP & AP50\% & Runtime (s) \\
    \hline
    16 & 1 & 11.80 & 35.04 & 0.039\\
    16 & 2 & 10.24 & 31.40 & 0.040\\
    16 & 4 & 11.59 & 34.11 & 0.040\\
    16 & 16 & \textbf{12.22} & \textbf{35.47} & 0.040\\
    \hline
    \end{tabular}
\end{table}

\begin{figure*}
    \centering
    \subfloat{\includegraphics[width=0.33\textwidth]{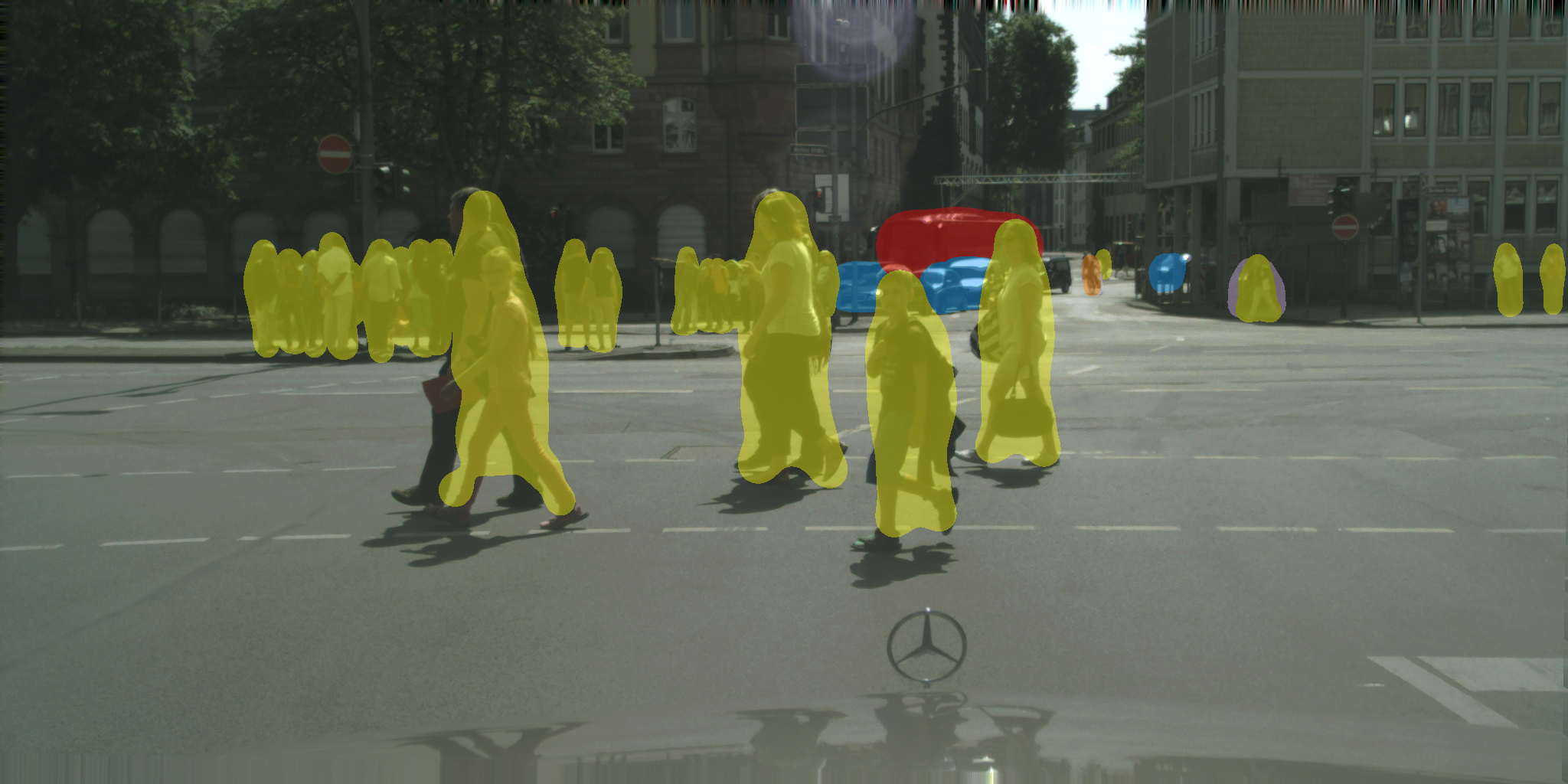}}
    \hfill
    \subfloat{\includegraphics[width=0.33\textwidth]{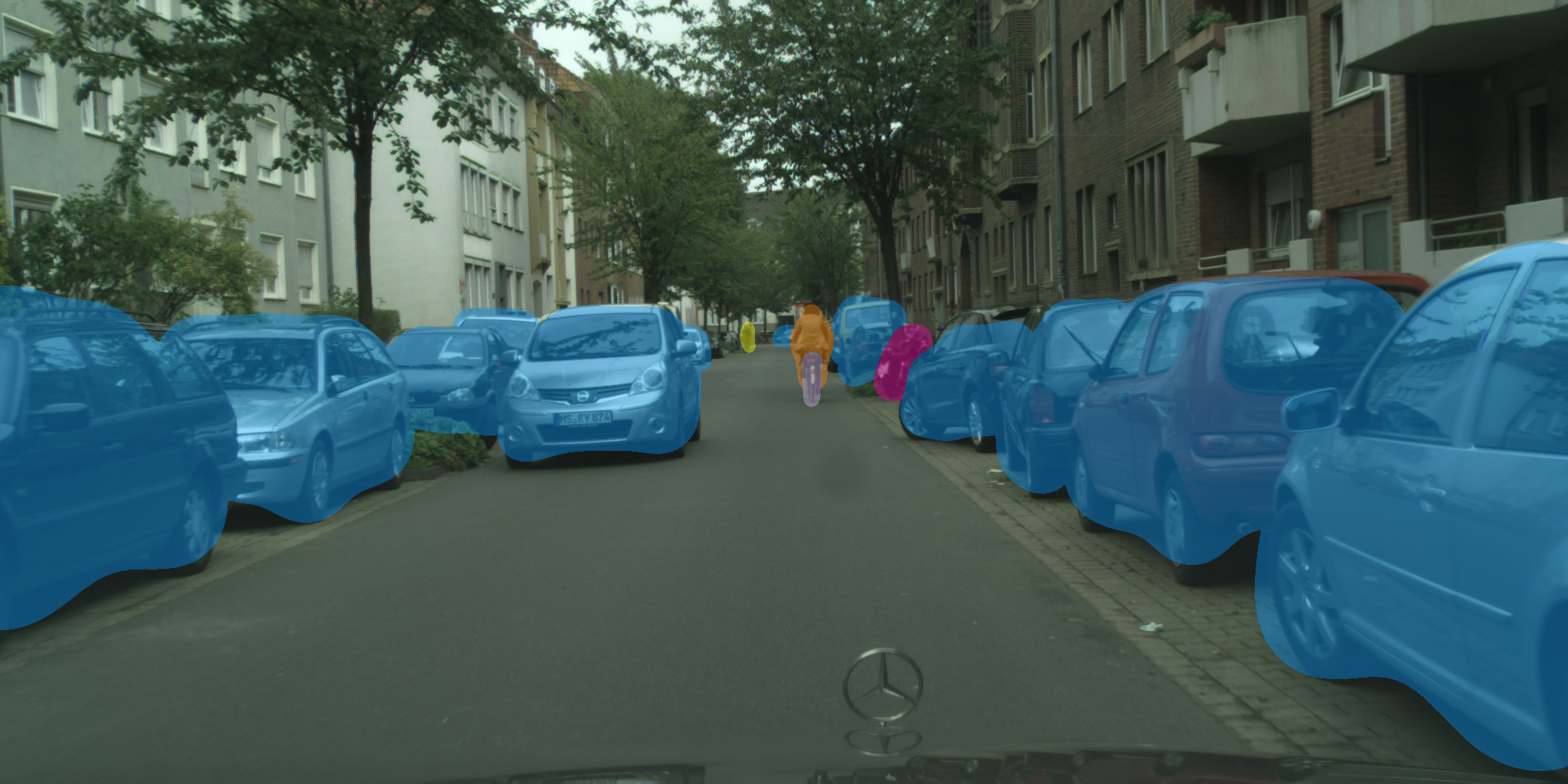}}
    \hfill
    \subfloat{\includegraphics[width=0.33\textwidth]{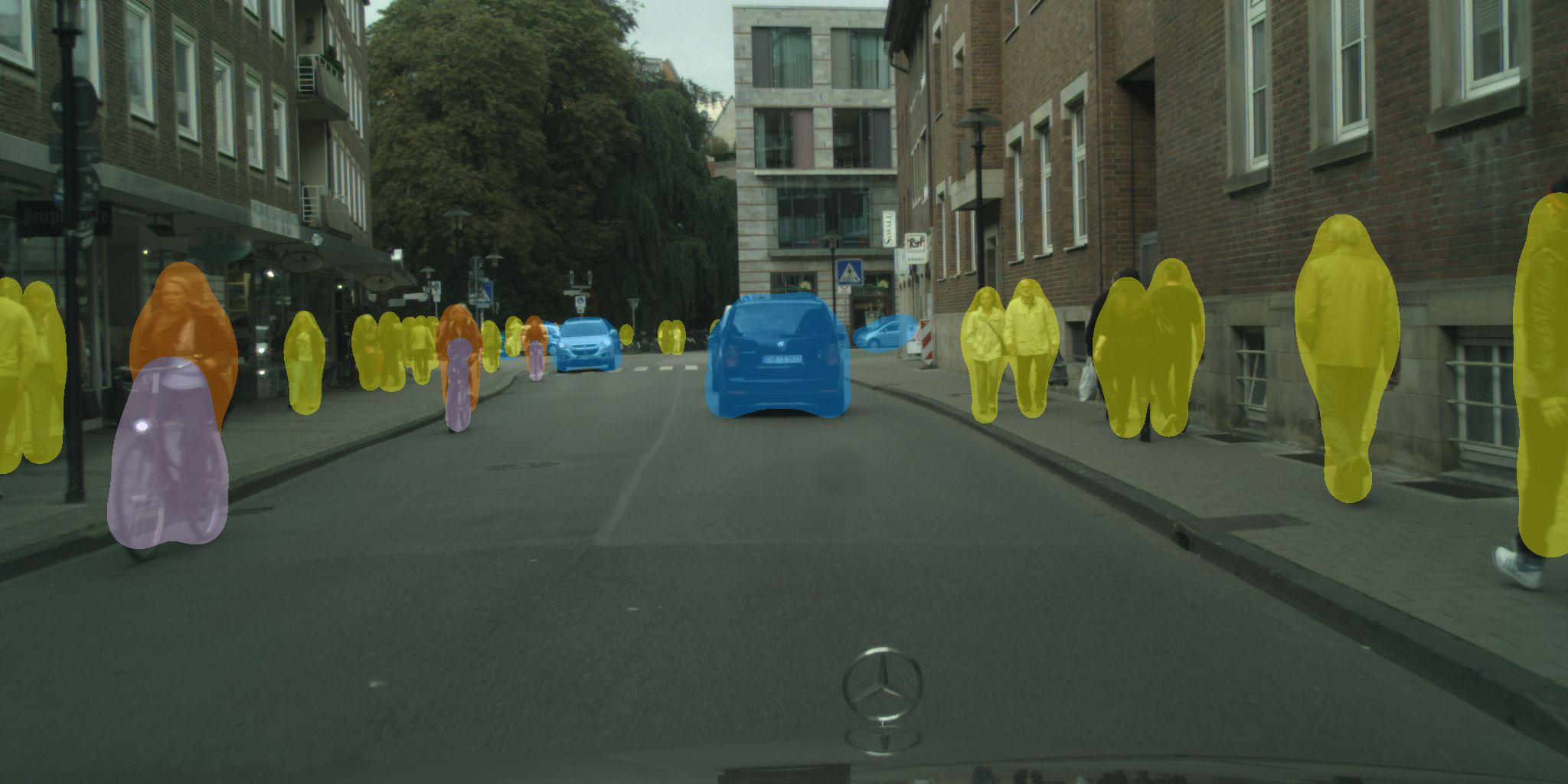}}
    \hfill
    \subfloat{\includegraphics[width=0.33\textwidth]{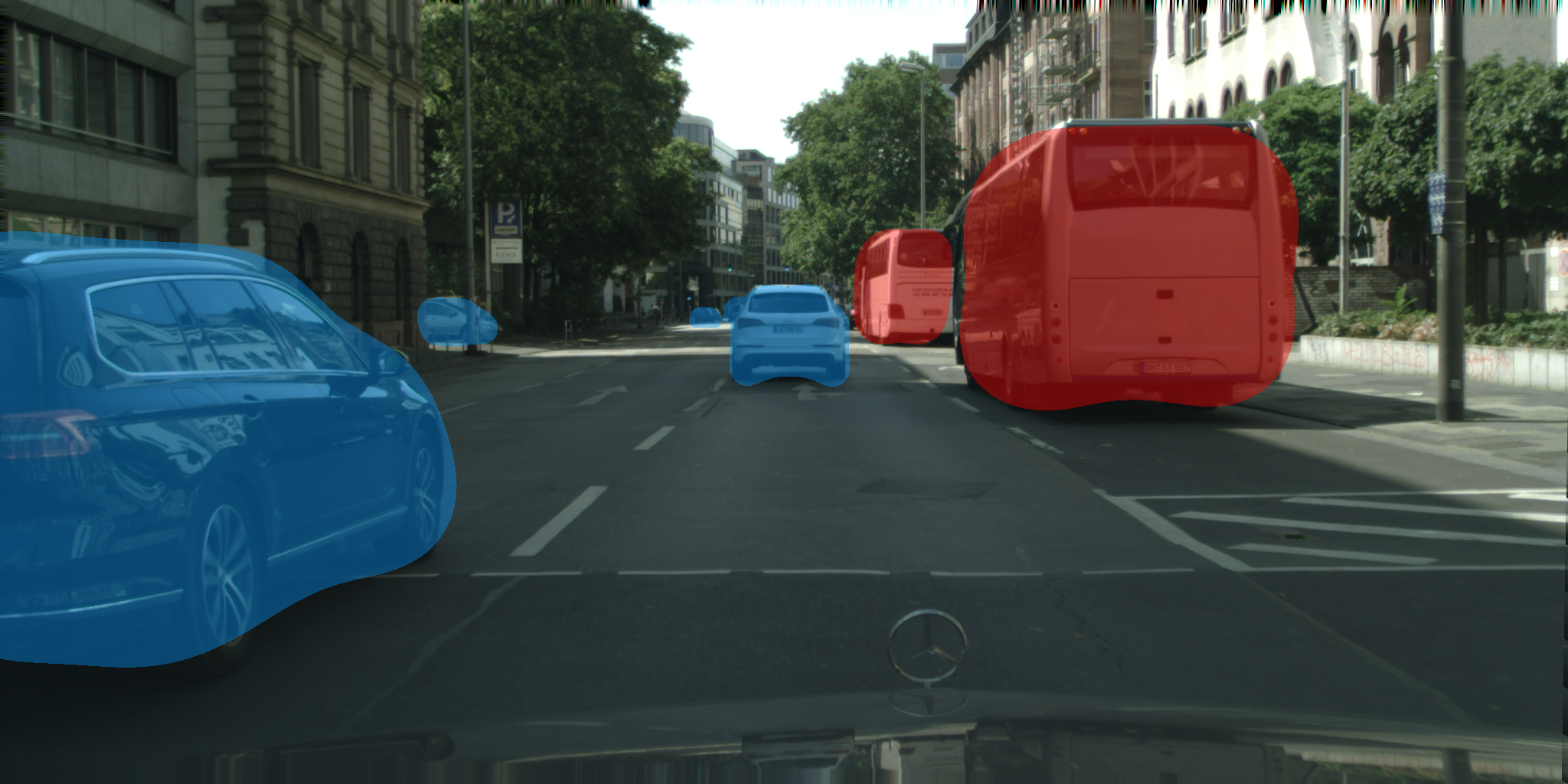}}
    \hfill
    \subfloat{\includegraphics[width=0.33\textwidth]{frankfurt_000001_035864_leftImg8bit.png}}
    \hfill
    \subfloat{\includegraphics[width=0.33\textwidth]{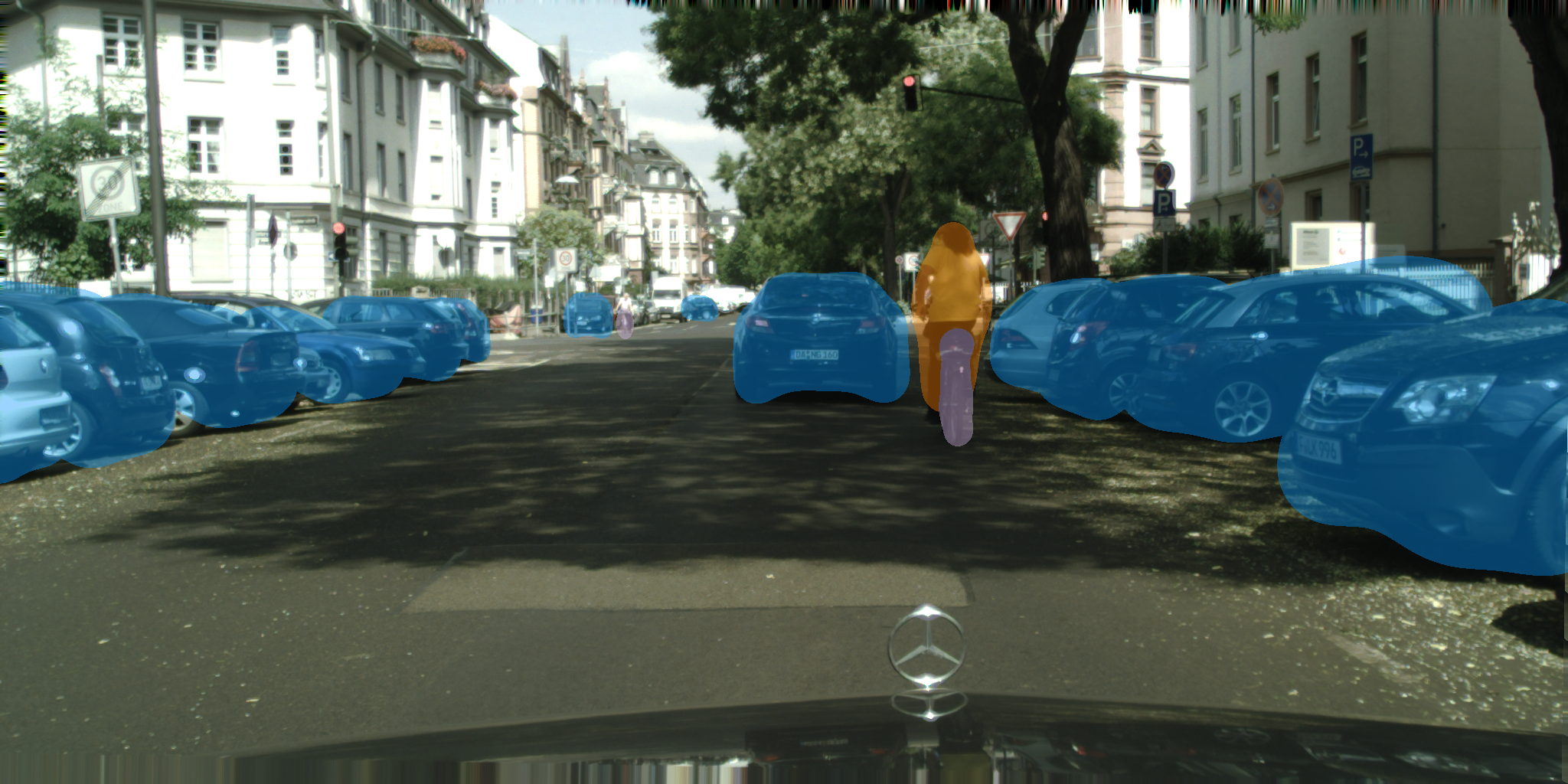}}
    \caption{Qualitative results on the Cityscapes test set. Colors correspond to semantic categories (blue - cars, yellow - person, red - bus, light pink - bicycle, orange - rider, violet - truck, bright pink - motorcycle). Best viewed on a screen.}
    \label{fig:qualitative_results}
\end{figure*}

The initial results of CenterDisks are somewhat round-shaped, which does not represent accurately the type of objects present in the datasets. Therefore, we studied the refinement of the masks during post-processing with the Douglas-Peucker algorithm (see section \ref{sec:postprocessing}).  With $\beta=0.01$, the AP is improved by 1.1 points, and the AP50\% is improved by 2.4 points (Table \ref{ablation_postprocessing}). We find that although this post-processing improves the final accuracy slightly, the inference time needed is not worth the improvement, with 1s per image. 

When using oracle prediction for the detection, the evaluation metrics reach the results of 12.42 for the AP and 39.35 for AP50\%. The accuracy is thus improved by 4 points for the AP50\%. It means that getting a better detection head could improve the results slightly.

 \begin{table}
    \renewcommand{\arraystretch}{1.3}
    \centering
    \caption{Ablation study for the post-processing step. The reduction factor for the Douglas-Peucker algorithm is defined in section \ref{sec:postprocessing}, and is proportional to $\beta$. Results on the Cityscapes validation set. We use the dice loss, with 16 different radii.
    \textbf{Boldface: Best results}.}
    \label{ablation_postprocessing}
    \begin{tabular}{llrrr}
    \hline
    N disks & $\beta$ & AP & AP50\% & Runtime (s) \\
    \hline
    16 & None & 12.22 & 35.47 & \textbf{0.040}\\
    16 & 0.001 & 13.03 & 37.52 & 0.925\\
    16 & 0.01 & \textbf{13.36} & \textbf{37.8}5 & 0.923\\
    \hline
    \end{tabular}
\end{table}

\subsection{Limitations}

The overall accuracy of our method does not reach the performance of the best existing instance segmentation methods. Due to the use of overlapping disks, it is almost impossible to get straight lines and sharp angles. This difficulty could only be overcome in using other shapes. Small objects are also hard to segment accurately, as we can see on the qualitative results (objects in the background in Figure \ref{fig:qualitative_results}) and when decomposing the results according to the category (Table \ref{fig:results_categories}). Finally, our method struggles also with fine separation, for example the legs of pedestrians. Nonetheless, our proposed representation can theoretically achieve a better mask than the polygonal methods for objects with holes, or whose center of gravity is outside the mask.

\section{Conclusion}

In this paper, we propose a new paradigm for mask approximation using disk covering. A fixed number of disks with different radii represent the objects. The model is trained without the need for elaborate or custom ground-truths. During training, the disks are projected as two-dimensional Gaussian functions on the image. It allows a direct comparison to the binary masks. This method shows promising results with regards to the accuracy-speed compromise. It could be further improved by refining the representation, with ellipses or more complex shapes. 
Finally, this segmentation approach could be generalized to other domains and could benefit to problems concerning more specifically the coverage of complex surfaces.

\IEEEtriggeratref{22}

\def\url#1{}
\bibliographystyle{IEEEtran}
\bibliography{IEEEabrv,Poly}

\end{document}